\documentclass[runningheads]{llncs}
\usepackage{graphicx}
\usepackage{tikz}
\usepackage{comment}
\usepackage{amsmath,amssymb} 
\usepackage{color}
\usepackage{epsfig}
\usepackage{graphicx}
\usepackage{amsmath}
\usepackage{amssymb}
\usepackage{cuted}
\usepackage{capt-of}
\usepackage{makecell}
\usepackage{booktabs}
\usepackage{multirow}
\usepackage{subfig}

\usepackage[accsupp]{axessibility}  

\usepackage[pagebackref,breaklinks,colorlinks]{hyperref}
\newcommand{\tablestyle}[2]{\setlength{\tabcolsep}{#1}\renewcommand{\arraystretch}{#2}\centering\footnotesize}

\begin{document}
\pagestyle{headings}
\mainmatter
\def\ECCVSubNumber{1710}  

\title{Generative Adversarial Network for Future Hand Segmentation from Egocentric Video} 

\titlerunning{Future Hand Segmentation}
\author{Wenqi Jia$^\star$\and
Miao Liu\thanks{Equal contribution.}\and
James M. Rehg}
\authorrunning{W. Jia et al.}
\institute{Georgia Institute of Technology, Atlanta, United States}
\maketitle

\begin{abstract}
We introduce the novel problem of anticipating a time series of future hand masks from egocentric video. A key challenge is to model the stochasticity of future head motions, which globally impact the head-worn camera video analysis. To this end, we propose a novel deep generative model -- EgoGAN. Our model first utilizes a 3D Fully Convolutional Network to learn a spatio-temporal video representation for pixel-wise visual anticipation. It then generates future head motion using the Generative Adversarial Network (GAN), and predicts the future hand masks based on both the encoded video representation and the generated future head motion. We evaluate our method on both the EPIC-Kitchens and the EGTEA Gaze+ datasets. We conduct detailed ablation studies to validate the design choices of our approach. Furthermore, we compare our method with previous state-of-the-art methods on future image segmentation and provide extensive analysis to show that our method can more accurately predict future hand masks. Project page: \href{https://vjwq.github.io/EgoGAN/}{https://vjwq.github.io/EgoGAN/}
\end{abstract}

\section{Introduction}

The egocentric vision paradigm provides an ideal vehicle for studying the relationship between visual anticipation and intentional motor behaviors, as head-worn cameras can capture both human visual experience and related sensory-motor signals. While prior works have recently addressed action anticipation in an egocentric setting~\cite{furnari2019rulstm,Ke_2019_CVPR,liu2019forecasting,shen2018egocentric,girdhar2021anticipative}, the problem of forecasting the detailed shape of hand movements in egocentric video remains unexplored. This is a significant deficit because many everyday motor behaviors cannot be easily categorized into specific action classes and yet play an important role in preparing and executing our routine activities. Such a general prediction capability could enable new applications in Augmented Reality (AR) and robotics, such as monitoring for safety in dangerous environments such as construction sites, or facilitating human-robot collaboration via improved anticipation.

\begin{figure*}[t]
\centering
\includegraphics[width=0.805\linewidth]{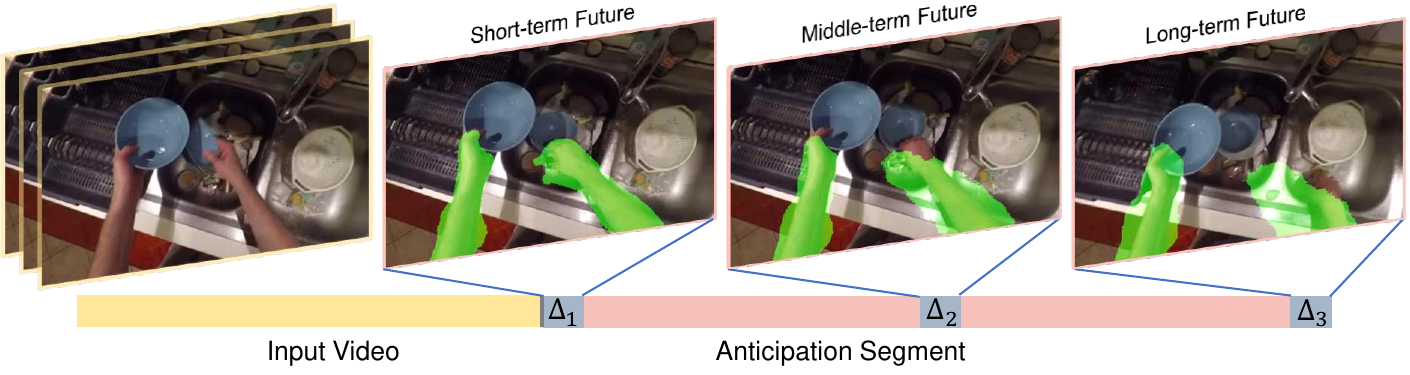}
\caption{\textit{Future hand segmentation task}: Given an input egocentric video, our goal is to predict a time series of future hand masks in the anticipation video segment. $\Delta_1$, $\Delta_2$, and $\Delta_3$ represent the short-term, middle-term, and long-term time points in the anticipation segment, respectively. The entanglement between drastic head motion and non-rigid hand movements poses a significant technical barrier in computer vision. Here, we visualize our forecasting results on this challenging task (best viewed in color).}
\label{fig:teaser}
\end{figure*}

To bridge this gap, this paper introduces a novel task of forecasting the detailed representation of future hand movements in egocentric video. Specifically, given an egocentric video, we seek to predict the hand masks of future video frames at three time points defined as short-term, middle-term, and long-term future  (see Fig.~\ref{fig:teaser} for a visual illustration of our problem setting). This task is extremely challenging for two reasons: 1) hands are deformable and capable of fast movement, and 2) head and hand motion are entangled in the egocentric video. Addressing these challenges requires the ability to 1) address the inherent uncertainty in anticipating the no-rigid hand movements, and 2) explicitly model the coordination between head and hand~\cite{pelz2001coordination}.

We attack the unique challenges of hand segmentation prediction by introducing a novel deep model -- \emph{EgoGAN}. Our model adopts a 3D Fully Convolutional Network (3DFCN) as the backbone to learn spatio-temporal video features. We then utilize the Generative Adversarial Network (GAN) to aid pixel-wise visual anticipation. Instead of using GAN to directly generate future video frame pixels from egocentric videos as in~\cite{zhang2017deep}, our key insight is to use the GAN to model an underlying distribution of possible future head motion. The adopted generative adversarial training schema can account for the uncertainty of future hand movements anticipation. In addition, the generated future head motion provides ancillary cues that complement video features for anticipating complex egocentric hand movements. Our end-to-end trainable EgoGAN model uses future hand masks as supervisory signals to train the segmentation network and estimated sparse optical flow maps from head motions to train the Generator and the Discriminator. At inference time, our model predicts a time series of future hand masks based \emph{only} on the egocentric video frame inputs.

To demonstrate the benefits of our proposed EgoGAN, we evaluate our model on two egocentric video datasets: EPIC-Kitchens 55~\cite{Damen2018EPICKITCHENS} and EGTEA Gaze+~\cite{Li_2018_ECCV}. We first conduct detailed ablation studies to validate our model design, and then compare our approach to the state-of-the-art methods on future image segmentation, demonstrating consistent performance gains on both datasets. We further provide visualizations to show the effect of our method. In summary, our paper makes following contributions:
\begin{itemize}
    \item We introduce a novel problem of predicting a time series of future hand masks from egocentric videos. 

    \item We propose a novel deep generative model -- EgoGAN, that hallucinates future head motions and further predicts future hand masks. To the best of our knowledge, we are the first to use a GAN to generate egocentric motion cues for visual anticipation.

    \item We conduct comprehensive experiments on two benchmark egocentric video datasets: EPIC-Kitchens 55~\cite{Damen2018EPICKITCHENS} and EGTEA Gaze+~\cite{li2020eye}. Our model achieves $1.3\%$ performance improvements on EPIC-Kitchens and $0.7\%$ on EGTEA in average F1 score. We also provide visualizations of our results and additional discussion of our method.
    

\end{itemize}

\section{Related Work}
We first review the most relevant works on egocentric vision. We then discuss previous literature on future image segmentation. Furthermore, we describe the related efforts on developing generative models for visual anticipation.

\noindent\textbf{Hands in Egocentric Vision}.\ Previous efforts on egocentric vision addressed a variety interesting problems, including action analysis~\cite{poleg2016compact,fathi2011understanding,Li_2018_ECCV,liu2021egocentric,Poleg_2014_CVPR,furnari2019rulstm,liu2019forecasting,soran2015generating,Ke_2019_CVPR,moltisanti2017trespassing} and social interaction understanding~\cite{fathi2012social,soo2015social,yonetani2016recognizing,yagi2018future}, etc. Here, we focus on discussing prior works on learning hand representations from egocentric videos. The most relevant work is from Liu et al.~\cite{liu2019forecasting}, where they factorized the future hand positions a latent attentional representation for action anticipation without considering the head motion. Similarly, Dessalene et al.~\cite{9340014} focused on predicting the hand-object interaction region of an action. Fathi et al.~\cite{fathi2012learning} utilized hand-eye coordination to design a probabilistic model for gaze estimation. Li et al.~\cite{li2015delving} showed how the motion patterns of the hands can be utilized for egocentric action recognition. Ma et al.~\cite{ma2016going} made use of a hand segmentation network to predict hand masks for localizing the object of interest and further recognizing the action. Shen et al.~\cite{shen2018egocentric} proposed to use hand mask and gaze fixation as additional cues for action anticipation. Rather than anticipating the hand movements, these previous works mainly use egocentric hand movements as an additional modality or intermediate representation for egocentric action understanding. Recently, Cai et al.~\cite{Cai_2020_CVPR} proposed a Bayesian-based domain adaptation framework for hand segmentation on egocentric video frames. In contrast, we address the novel task of predicting pixel-wise hand masks, which captures the fine-grained details of future hand movements.


\noindent\textbf{Future Segmentation}.\
A rich set of literature addressed the related but vastly different task of video segmentation~\cite{yang2019video,tsai2016video,xu2018dynamic,chandra2018deep,nilsson2018semantic}. We refer to a recent survey~\cite{wang2021survey} for a thorough discussion on this topic. Note that previous works on video segmentation seek to track the instance masks within the video segment, and therefore do not apply to the anticipation setting, where the information of future video frames is not accessible for making an inference. Fewer works address the more relevant topic of future image semantic segmentation. Luc et al.~\cite{luc2017predicting} first investigated the problem of semantic segmentation of future video frames and further extended their work to future instance segmentation~\cite{luc2018predicting}. Nabavi et al.~\cite{rochan2018future} utilized the ConvLSTM network to model the temporal correlations of video sequences for future semantic segmentation. Jin et al.~\cite{jin2017predicting} proposed to anticipate the future optical flow and future scene segmentation jointly. Recently, Chiu et al.~\cite{chiu2020segmenting} introduced a teacher-student knowledge distillation model for predicting the future semantic segmentation based on preceding video frames. Building on these prior works, we propose the first model to address the future segmentation problem under the challenging egocentric setting. It is worth noting that previous methods recursively predict future segmentation, in which the current anticipation result is used as the input for predicting the segmentation of the next time step. In contrast, we use a 3D Fully Convolutional Network (3DFCN) to predict a time series of future hand masks in one shot. In Sec.~\ref{sec:results}, we show that the 3DFCN can effectively capture the spatio-temporal video features for pixel-wise visual anticipation in an end-to-end fashion. We also compare our EgoGAN model to those relevant works and demonstrate a clear performance improvement.

\noindent\textbf{Generative Models for Visual Anticipation}.\
Tremendous efforts have been made in action anticipation~\cite{kitani2012activity,vondrick2016anticipating,gao2017red,kataoka2016recognition,furnari2019rulstm,liu2019forecasting,soran2015generating,Ke_2019_CVPR,rodriguez2018action,guan2020generative} and generative adversarial networks~\cite{gregor2015draw,odena2017conditional,zhang2017stackgan,goodfellow2014generative,mirza2014conditional,isola2017image}. Here we mainly discuss previous investigations on forecasting the human body motions using generative models. Fragkiadaki et al.~\cite{fragkiadaki2015recurrent} proposed to use a recurrent network for predicting and generating the human body poses and dynamics from videos. A similar idea was also explored in~\cite{gui2018adversarial}. Walker et al.~\cite{walker2016uncertain} utilized Variational Autoencoders (VAE) for predicting the dense trajectories of video pixels. They further leveraged human body poses as an intermediate feature for generating future video frames with a Generative Adversarial Network (GAN)~\cite{walker2017pose}. Gupta et al.~\cite{gupta2018social} explored a GAN-based model for forecasting human trajectories. Zhang et al.~\cite{zhang2020generating,zhang2021we} developed a Conditional Variational Autoencoder to generate human body meshes and motions in 3D scenes. Despite the success in forecasting body motion, the use of GANs was largely understudied in egocentric vision. Zhang et al.~\cite{zhang2017deep} used a GAN to generate future video frames and further predict future gaze fixation. Though GAN has the capability of addressing the uncertainty of data distribution, using GANs to directly forecast pixels in video~\cite{tulyakov2018mocogan} remains a challenge, especially when there exists drastic background motion in the egocentric videos~\cite{zhang2017deep}. In contrast, our method adopts the adversarial training mechanism to model the underlying distribution of possible future head motion, and thereby captures the drastic change of scene context in egocentric video. In the ablation study, we show that our approach outperforms a baseline model that uses GAN to directly predict future hand masks.

\begin{figure*}[t]
\centering
\includegraphics[width=1.0\linewidth]{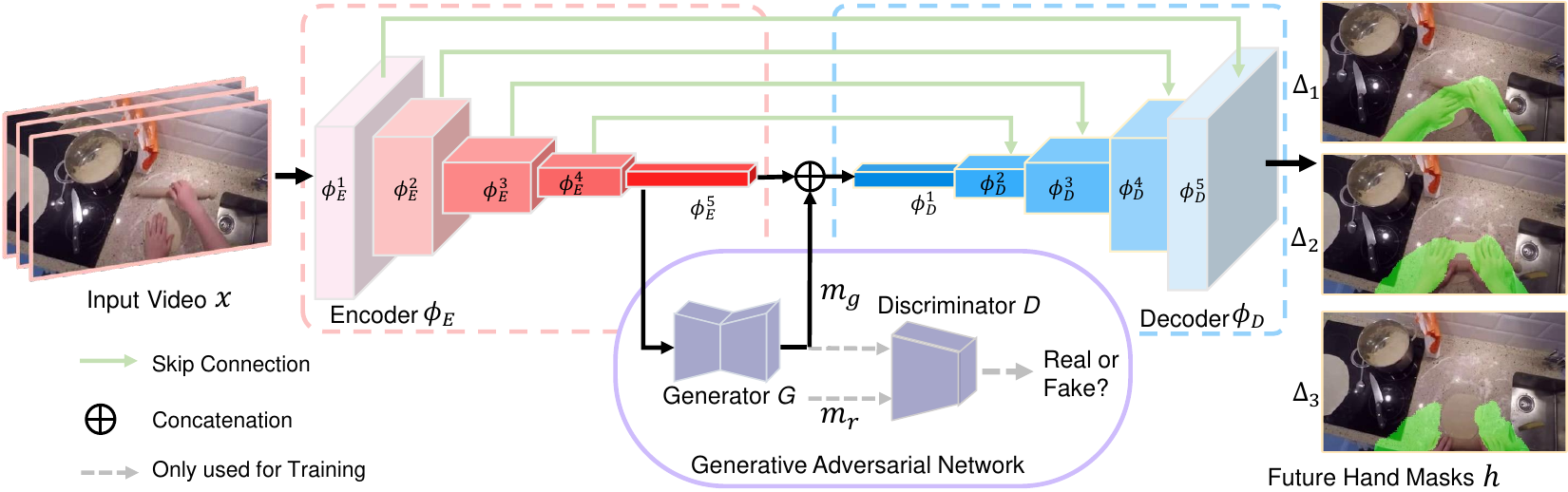}
\captionof{figure}{\textit{Overview of our proposed \textit{EgoGAN} model}. Our model takes egocentric video frames as the inputs, and outputs future hand masks at different time steps. It is composed of a {\fontseries{b}\selectfont 3D Fully Convolutional Network (3DFCN)} and a {\fontseries{b}\selectfont Generative Adversarial Network (GAN)}. The Encoder Network $\phi_E$ in the \textbf{3DFCN} extracts video features from the input frames, and is then separated into two branches: (1) encoded feature $\phi_E(x)$ is fed into the {\fontseries{b}\selectfont Generator (G)} in {\fontseries{b}\selectfont GAN} for generating fake future head motion $m_g$, and a {\fontseries{b}\selectfont Discriminator (D)} is trained to distinguish the generated future head motion from the real ones; (2) $m_g$ is concatenated to $\phi_E(x)$ and the concatenated tensor are then fed into the Decoder Network $\phi_D$ in {\fontseries{b}\selectfont 3DFCN}. Finally, the encoder features are further combined with corresponding decoder features using skip connections for future hand mask prediction.}
\label{fig:overview}
\end{figure*}

\section{Method}
\label{sec:method}
Given an input egocentric video $x=\{x^1, ..., x^t\}$, where $x^t$ is the video frame indexed by time $t$, our goal is to predict a time series of future hand masks $h=\{h^{t+\Delta_1}, h^{t+\Delta_2}, h^{t+\Delta_3}\}$. As illustrated in Fig.~\ref{fig:teaser}, we consider hand segmentation as a binary classification problem: the value of $h^{i}(x,y)$ can be viewed as the probability of spatial position $(x,y)$ being a hand pixel at time step $i$, where $i \in \{t+\Delta_1, t+\Delta_2, t+\Delta_3\}$. $\Delta_1$, $\Delta_2$, and $\Delta_3$ represent the time steps for short-term, middle-term, and long-term future segmentation, respectively. This three-steps-ahead visual anticipation setting is also used in previous works on future image segmentation~\cite{luc2017predicting,rochan2018future}.

We now present an overview of our \textbf{EgoGAN} model in Fig.~\ref{fig:overview}. We make use of a 3D Fully Convolutional Network (3DFCN) $\phi$ as the backbone model for future hand segmentation. The 3DFCN is composed of a 3D convolutional encoder $\phi_E$ and a 3D deconvolutional decoder $\phi_D$. We further adopt a Generative Adversarial Network (GAN) for learning future head motions. Specifically, a Generator network (G), composed of 3D convolutional operations, is used to generate future head motion $m_g$ based on the encoded video feature $\phi_E(x)$. A Discriminator Network (D) is trained to distinguish the fake future head motions $m_g$ from real future head motions $m_r$. Finally, $\phi_D$ combines $m_g$ and $\phi_E(x)$ for predicting future hand masks. In the following sections, we detail each key component of our model.

\subsection{3D Fully Convolutional Network}
We first introduce the 3D Fully Convolutional Network (3DFCN) backbone in our method. We use an I3D model~\cite{wang2018non} as the backbone encoder network $\phi_E$ for learning spatio-temporal video representations. Following~\cite{simonyan2014very,he2016deep}, $\phi_E$ has $5$ convolutional blocks, thereby producing video features at different spatial and temporal resolutions. Following~\cite{long2015fully}, we construct the decoder network $\phi_D$ symmetric to $\phi_E$. Therefore, $\phi_D$ is also composed of $5$ deconvolution layers. We denote the encoder and decoder video features from the $i$th convolutional block as $\phi^i_E(x)$ and $\phi^i_D(x)$, respectively (See Fig.~\ref{fig:overview} for the index naming of $\phi_E$ and $\phi_D$). The features of each decoder layer are combined with the features from the corresponding encoder block with skip connections and are then fed into the next layer. Formally, we have:
\begin{equation}
\phi^{i+1}_D(x) = deconv(\phi^{i}_D(x)+\phi^{6-i}_E(x)),
\label{eq:decoder}
\end{equation}
where $i\in \{1,2,3,4\}$. We design our decoder so that $\phi^{i}_D$ produces a feature map with the same tensor size as $\phi^{6-i}_E(x)$. The deconvolution operation is implemented with 3D transposed convolution. Note that the last deconvolution layer of $\phi_D$ produces a tensor of the same size as the input video ($T\times W \times H$). We further apply a 3D convolutional operation with a kernel size of $k\times1\times1$ to predict the future hand mask tensor $h$ with size $3\times W\times H$, where each temporal slice corresponds to the predicted hand masks of the short-term, middle-term, and long-term future video frames. We describe the details of our network architecture in the supplementary materials.

\subsection{Generative Adversarial Network}
\label{sec:methodgan}
The key to our approach is to use the Generative Adversarial Network (GAN) to hallucinate the future head motions for future hand mask segmentation. Our design choice stems from the observation that head motion causes drastic changes in the active object cues and background scene context captured in the egocentric videos, and this motion is closely related to hand movements. Therefore, we seek to explicitly encode the future head motion cues for hand motion anticipation. Moreover, visual anticipation has intrinsic ambiguity -- similar current observations may correspond to different future outcomes. This observation motivates us to use the adversarial training scheme to account for the inherent uncertainty of future representation. In this section, we introduce the egocentric head motion representation. We then describe the design choice and learning objective of the GAN in our method. 

\noindent\textbf{Egocentric Head Motion Representation}.\ In the egocentric setting, head motion is implicitly incorporated in the video itself. Thus, we follow~\cite{li2013learning} to use the sparsely sampled optical flow to represent the egocentric head motion. As mentioned before, the real future head motion is denoted as $m_r$, and is only available for training.

\noindent\textbf{Generator Network and Discriminator Network}.\ The generator network (G) takes video feature $\phi_E(x)$ as inputs and generates future head motions $m_g = G(\phi_E(x))$. Following~\cite{vondrick2016generating,zhang2017deep,isola2017image,walker2017pose}, G does not take any noise variables as additional inputs. This is because the $\phi_E(x)$ is a latent representation that incorporates the noisy signals of visual anticipation. G is composed of multiple 3D convolutional operations and a nonlinearity function, and is trained to produce a realistic $m_g$ that is difficult to distinguish from $m_r$ for an adversarially-trained discriminator network (D). D takes future head motion samples as inputs and determines whether the input sample is real or fake. It is composed of 3D convolutional operations and a sigmoid function for binary classification, and is trained to classify the input sample as either real or generated.

\noindent\textbf{Learning Objective of GAN}.\ We now formally define the objective function of the GAN in our method. The objective function for training the discriminator network is given by:
\begin{equation}
\mathcal{L}_d = \mathcal{L}_{ce}(D(m_r),1) + \mathcal{L}_{ce}(D(m_g),0)
\label{eq:ld},
\end{equation}
where $\mathcal{L}_{ce}$ is the standard cross-entropy loss for binary classification. The generator loss $\mathcal{L}_{g}$ can be formulated as:
\begin{equation}
\mathcal{L}_g = \mathcal{L}_{ce}(D(m_g),1)+\lambda \vert m_g - m_r\vert
\label{eq:lg}.
\end{equation}
Here, we follow~\cite{pathak2016context} to adopt a traditional L1 distance loss that encourages the generated sample to be visually consistent with the real sample, while $\lambda$ denotes the weight to balance the two loss terms.

\subsection{Full Model of EgoGAN}
We now summarize the full architecture of our proposed EgoGAN model. The main idea is to explicitly model the underlying distribution of possible future head motion $m_g$ with the GAN, and use $m_g$ as additional cues to facilitate future hand mask segmentation from the video representations of the encoder network. Specifically, the video feature from the last encoder block $\phi^{5}_E(x)$ and generated future head motions $m_g$ are concatenated and fed into the first layer of the decoder as inputs. Therefore, we have:
\begin{equation}
\phi^{1}_D(x) = deconv(\phi^{5}_E(x)\oplus m_g).
\label{eq:fullmodel}
\end{equation}
Hence, the decoder network jointly considers $\phi_E(x)$ and $m_g$ for predicting future hand masks $h$.

\noindent\textbf{Training and Inference}.\  We use the binary cross-entropy loss to train the 3DFCN encoder and decoder: 
\begin{equation}
\mathcal{L}_{seg} = \mathcal{L}_{ce}(\phi_D(\phi_E(x), m_g), \hat{h}),
\label{eq:loss_seg}
\end{equation}
where $\hat{h}$ denotes the ground truth of future hand masks. We adopt the standard adversarial training pipeline in~\cite{goodfellow2014generative}, where G and D are trained to play against each other. Therefore, we let the gradients alternatively flow through D, and then G. Moreover, we freeze the encoder weights during the gradient step on G and D, and freeze the generator weights during the gradient step on the 3DFCN to isolate their training processes from each other. 

Note that our model does not need the real future head motion as additional inputs at inference time. Instead, our model can generate future head motion and further predict future hand masks based on only raw video frames.

\subsection{Implementation Details}

\noindent\textbf{Network Architecture}.\ We adopt an I3D-Res50 model~\cite{carreira2017quo,wang2018non} that is pre-trained on Kinetics as the backbone encoder network. It is composed of five 3D convolutional blocks, connecting with a symmetrical decoder network that contains five 3D deconvolutional layers. As for the GAN network, the generator network takes the video features from the $5$th block of the encoder network as inputs and produces a low-resolution future head motion flow map as output. The discriminator network serves as a binary classifier to supervise the quality of the output of the generator. Our model is implemented in PyTorch and will be made publicly available.

\noindent\textbf{Training Schema}.\ As discussed in the previous section, the gradients step separately for 3D Fully Convolutional Network (3DFCN), Generator (G), and Discriminator (D). The 3DFCN model is trained using an SGD optimizer with momentum of 0.9. The initial learning rate is 0.1 with cosine decay. We set weight decay to 1e-4 and enable batch norm~\cite{ioffe2009batch}. G and D are trained using the Adam Optimizer with momentum parameters $\beta_1=0.5$, $\beta_2 = 0.999$ and a initial learning rate of 0.01 with cosine decay. Our model was trained for $70$ epochs with batch size 16 on 4 GPUs, and synchronized batch normalization was enabled. 

\noindent\textbf{Data Processing}.\ 
We downsampled all video frames to a height of $256$ while preserving the original aspect ratio. For training, we applied several data augmentation techniques, including random flipping, rotation, cropping, and color jittering to avoid overfitting. Our model takes an input of 8 frames (temporally sampled by 8) with a resolution of $224\times224$. We use the TV-L1 algorithm~\cite{perez2013tv} to compute the optical flow, and sparsely sample from the computed flow map to approximate the head motion as discussed in Sec.~\ref{sec:methodgan}. Therefore, the head motion is represented as a sparse flow map spatially downsampled by 32. At inference time, our model takes the downsampled videos with the original aspect ratio as inputs, and predicts the future hand masks.

\section{Experiments}

\subsection{Dataset and Metrics}
\label{sec:data}
\noindent\textbf{Dataset}.\ We make use of two egocentric video benchmark datasets:  EPIC-Kitchens 55~\cite{Damen2018EPICKITCHENS} and EGTEA Gaze+~\cite{li2020eye}. For the EPIC-Kitchens dataset, we set $\delta_{1,2,3}=\{1,15,30\}$, which corresponds to a long-term anticipation time of $1.0s$. As for the  EGTEA dataset, we set $\Delta_{1,2,3}=\{1,6,12\}$, which corresponds to an anticipation time of $0.5s$, because EGTEA has a smaller angle of view in comparison with the EPIC-Kitchens. The same anticipation time setup is also adopted in~\cite{liu2019forecasting}. To encourage our model to capture the meaningful preparation and planning process of daily actions, we segment the data so that the long-term future frame is chosen right before the beginning of each trimmed action segment annotated in EPIC-Kitchens and EGTEA. We use the train/val split provided by~\cite{furnari2019rulstm} for EPIC-Kitchens 55 and the train/test split1 from EGTEA. We remove the instances where hands are not captured within the anticipation segment, which results in 11,935/2,746 (train/val) samples on EPIC-Kitchens, and 4,042/991 (train/test) samples on EGTEA. 

\noindent\textbf{Hand Mask Ground Truth}.\ For the EPIC-Kitchens dataset, we use the domain adaption method introduced in~\cite{Cai_2020_CVPR} to generate the ground truth hand masks. ~\cite{Cai_2020_CVPR} has empirically verified the quality of generated hand masks. As for the EGTEA dataset, we train a 2D FCN model for frame-level hand segmentation using the provided hand mask annotation. As discussed in~\cite{li2017learning}, the FCN model can generalize well on the entire dataset. We thus use the inference results on the anticipation video frames as the ground truth of future hand masks. 

\noindent\textbf{Metrics}.\ As discussed in Sec.~\ref{sec:method}, we consider future hand segmentation as a pixel-wise binary classification problem. Previous future image segmentation works~\cite{chiu2020segmenting,jin2017predicting} use pixel accuracy and mIoU as evaluation metrics. However, pixel accuracy does not penalize the false-negative prediction of the long-tailed distribution, and mIoU can not properly evaluate the shape of the predicted masks for binary segmentation. Therefore, we follow~\cite{Li_2018_ECCV,liu2019forecasting} to report Precision and Recall values together with their corresponding F1 scores.

\subsection{Model Ablations and Analysis}
\label{sec:ablation}
To validate our model design, we conduct experiments on ablations and variations in our model. Specifically, we investigate how the egocentric head motion cues facilitate future hand segmentation and demonstrate the benefits of using the GAN for modeling future head motion. We also show how modeling the future gaze as attentional representation affects the future hand segmentation performance.

\begin{table}[t]
\centering
\caption{\textit{Analysis of variations in our approach}. We conduct detailed ablation studies to validate our model design, and further show the results of variations of our method to demonstrate the benefits of using the GAN for modeling future head motion. *: HeadDir takes future head motions as additional input modalities at inference time, which in fact violates the future anticipation setting (See more discussion in Sec.~\ref{sec:ablation}). The best results are highlighted with \textbf{boldface}.}
\subfloat[Experimental Results on EPIC-Kitchens Dataset]{
\tablestyle{4pt}{1.0}
\scalebox{0.84}{
\begin{tabular}{c|ccc}
\hline
\multicolumn{1}{c|}{\multirow{2}{*}{Method}}          
 &\multicolumn{3}{c}{EPIC-Kitchens (Precision/ Recall/ F1 Score)} \\ \cline{2-4}
\multicolumn{1}{c|}{}   & short-term & middle-term  & long-term     \\ \hline
\makecell{Future Gaze}        & N/A & N/A  & N/A\\
\makecell{HeadDir* }  & 70.55/ 71.33/ 70.94 & 43.15/ 53.66/ 47.83 & 30.51/ 49.60/ 37.78 \\
\hline
\makecell{3DFCN (w/o GAN, w/o Head)} & 69.51/ 70.81/ 70.15 & 42.51/ 51.66/ 46.64 & 29.88/ 47.46/ 36.67 \\
\makecell{HeadReg (w/o GAN, w/ Head)}  & 70.46/ 70.25/ 70.36  & 41.41/ 52.55/   46.32 & 29.22/ 48.50/ 36.47 \\

\makecell{DirectGan (w/ GAN, w/o Head)}  & 69.12/ {\fontseries{b}\selectfont 71.60}/ 70.34  & 43.83/ 51.32/ 47.28   & 30.76/ 47.48/ 37.33  \\
\makecell{EgoGAN (w/ GAN, w/ Head)} & {\fontseries{b}\selectfont 70.89}/ 71.24/ {\fontseries{b}\selectfont 71.07} & {\fontseries{b}\selectfont 43.79}/ {\fontseries{b}\selectfont 53.23}/ {\fontseries{b}\selectfont 48.05} & {\fontseries{b}\selectfont 31.39}/ {\fontseries{b}\selectfont 48.57}/ {\fontseries{b}\selectfont 38.14}\\ 
\hline
\end{tabular}}
} \\
\subfloat[Experimental Results on EGTEA Gaze+ Dataset]{
\tablestyle{4pt}{1.0}
\scalebox{0.84}{
\begin{tabular}{c|ccc}
\hline
\multicolumn{1}{c|}{\multirow{2}{*}{Method}} &\multicolumn{3}{c}{EGTEA (Precision/ Recall/ F1 Score)} \\ 
\cline{2-4}
\multicolumn{1}{c|}{}   & short-term & middle-term  & long-term \\ 
\hline
\makecell{Future Gaze} & 45.17/ 59.94/ 51.51 & 38.63/ 64.02/ 48.19 & 35.71/ 63.78/ 45.78 \\
\makecell{HeadDir* }   & 44.58/ 63.87/ 52.51 & 41.29/ 60.65/ 49.13 & 39.36/ 59.02/ 47.23 \\
\hline
\makecell{3DFCN (w/o GAN, w/o Head)}   & 43.62/ {\fontseries{b}\selectfont 61.69}/ 51.11 &40.25/ 58.93/ 47.83 & 37.83/ 58.32/ 45.89\\
\makecell{HeadReg (w/o GAN, w/ Head)}  & 43.54/ 61.03/ 50.82 &{\fontseries{b}\selectfont 41.31}/ 55.24/ 47.27 & 36.87/ 58.23/ 45.15\\
\makecell{DirectGan (w/ GAN, w/o Head)} & 43.78/ 61.33/ 51.09 &38.38/ {\fontseries{b}\selectfont 63.81}/ 47.93   & 35.53/ {\fontseries{b}\selectfont 63.41}/ 45.54 \\
\makecell{EgoGAN (w/ GAN, w/ Head)}    &{\fontseries{b}\selectfont 44.91}/ 61.48/ {\fontseries{b}\selectfont 51.91} & 41.10/ 59.90/ {\fontseries{b}\selectfont 48.75} &{\fontseries{b}\selectfont 38.16}/ 59.88/ {\fontseries{b}\selectfont 46.61}\\ 
\hline
\end{tabular}}}
\label{table:ablation}
\end{table}


\noindent\textbf{Benefits of Encoding Future Head Motions}.\ As a starting point, we compare the model that uses only the 3D Fully Convolutional Network (denoted as \textit{3DFCN}) with the model that directly takes future head motion as an additional input modality (denoted as \textit{HeadDir}). HeadDir shares the same backbone network as 3DFCN, but requires the future head motions for making an inference and therefore violates the future anticipation setting, where the model can not use any information from the anticipation video segment for making an inference. HeadDir quantifies the performance improvement when the egocentric head motion cues are explicitly encoded into the model in a two-stream structure~\cite{simonyan2014two}. The experimental results are summarized in Table~\ref{table:ablation}. Compared to 3DFCN, HeadDir achieves a large performance gain on EPIC-Kitchens ($+0.8\%/1.2/1.1\%$ in F1 score for short/middle/long term anticipation), and reaches ($+1.4\%/1.3\%/1.3\%$) on EGTEA. 


Our method, on the other hand, consistently outperforms 3DFCN on both EPIC-Kitchens($+0.9\%/1.5\%/1.8\%$) and EGTEA ($+0.8\%/0.9\%/0.7\%$). More importantly, our method improves HeadDir by $+0.1\%/0.2\%/0.4\%$ on EPIC-Kitchens. This result suggests that the GAN from our model does not simply learn to predict a future head motion flow map; instead, it models the underlying distribution of possible future head motion and thus improves the future hand anticipation accuracy by addressing the inherent uncertainty of visual forecasting. It is to be observed that our model slightly lags behind HeadDir ($0.6\%/0.4\%/0.6\%\downarrow$) on EGTEA, because EGTEA has fewer samples to train our deep generative model. And we also re-emphasize that our method does not use any additional inputs at inference time as in HeadDir. \\
\noindent\textbf{The Effect of GAN}.\ To further show the benefits of using the GAN for learning future head motions, we consider a baseline model -- \textit{HeadReg}, that uses a regression network to predict future head motions with only L1 distance in Eq.~\ref{eq:lg}. Note that the regression network is implemented the same way as the generator network from EgoGAN. As shown in Table~\ref{table:ablation}, without using an adversarial training mechanism in our approach, HeadReg lags behind our model by $0.7\%/1.7\%/1.7\%\downarrow$ and $1.1\%/1.5\%/1.5\%\downarrow$ in F1 score for short/middle/long term anticipation on EPIC-Kitchens and EGTEA, respectively. These results support our claim that the GAN can address the stochastic nature of representation and thereby outperforms HeadReg by a notable margin on the future hand segmentation task. \\
\noindent\textbf{Video Pixel Generation vs. Head Motion Generation}.\ We denote another baseline model that directly uses a GAN for anticipating future hand masks, as \textit{DirectGan}. This model is composed of the 3DFCN backbone network that generates the future hand masks, and a discriminator network that classifies whether the given hand masks are real or not. The results are presented in Table~\ref{table:ablation}. Importantly, the adversarial training schema in DirectGan slightly decreases the performance of 3DFCN model on EGTEA, and has minor improvement on EPIC-Kitchens. We speculate that this is because directly using a GAN for predicting future hand masks cannot effectively capture the drastic change of scene context in egocentric video. In contrast, our model uses a GAN to explicitly model the head-hand coordination in the egocentric video thereby being capable of more accurately forecasting egocentric hand masks. \\
\noindent\textbf{Future Head Motion vs. Future Gaze}.\ Furthermore, we present experimental results on how modeling future gaze fixation affects future hand segmentation. Note that the gaze tracking data is only available for the EGTEA dataset. Specifically, we make use of a GAN to model the probabilistic distribution of future gaze fixation. Instead of concatenating future gaze with encoded video features as in Eq.~\ref{eq:fullmodel}, we follow~\cite{Li_2018_ECCV} to use gaze distribution as a saliency map to select important spatio-temporal video features with element-wise multiplication. As shown in Table~\ref{table:ablation}, the resulting future gaze model slightly outperforms the baseline 3DFCN model, yet lags behind our model that uses head motion as the key representation ($0.7\%/0.6\%/0.6\%\downarrow$ in F1 score on EGTEA). Previous work~\cite{li2013learning}  suggested that eye-head-hand coordination is important for egocentric gaze estimation, while our results further show that exploiting the eye-head-hand coordination is also beneficial for pixel-wise egocentric visual anticipation. Moreover, future head motion potentially plays a more important role than future gaze fixation on our fine-grained hand forecasting task. \\
\noindent\textbf{Analysis on Ablation Studies}.\
To help interpret the performance improvement of our method, we consider a baseline 3DFCN model that uses dense I3D-Res101 as the encoder network. Importantly, with 50 more layers, the I3D-Res101 backbone can only improve the model performance by $+0.1\%/0.3\%/0.3\%$ on EPIC-Kitchens and $+0.7\%/0.4\%/0.5\%$ on EGTEA. As shown in Table~\ref{table:ablation}, our model has a larger performance improvement than switching to a dense encoder network. In supplementary material, we also present additional results of our model using the I3D-Res101 backbone and further demonstrate our method is a robust approach that can generalize to different backbone networks.
\begin{table*}[t]
\centering
\caption{\textit{Comparison with previous state-of-the-art methods on future image segmentation}. Our results consistently outperform the second-best results (across all methods) by +1.3\% on EPIC-Kitchens and +0.7\% on EGTEA in average F1 score. *: We re-implement the model to take raw video frames as inputs as our method (See more discussion in Sec.~\ref{sec:results}). The best results are highlighted with \textbf{boldface}, and the second-best results are \underline{underlined}.}
{


\subfloat[Experimental Results on EPIC-Kitchens Dataset]{
\tablestyle{6pt}{1.0}
\scalebox{0.92}{
\begin{tabular}{c|ccc}
\hline
\multicolumn{1}{c|}{\multirow{2}{*}{Method}}
&\multicolumn{3}{c}{Epic-Kitchens (Precision/ Recall/ F1 Score)} \\ \cline{2-4}
\multicolumn{1}{c|}{}    & short-term & middle-term  & long-term     \\ \hline 
\makecell{X2X~\cite{luc2017predicting}}   &  68.69/ 69.35/ 69.02 & 40.81/ 50.61/ 45.18  & 28.14/ 45.76/ 34.85\\
\makecell{ConvLSTM~\cite{rochan2018future}} & 69.02/ 69.44/ 69.22 &42.72 /51.78/ 46.82  & 30.01/ 48.01/ \underline{36.94} \\
\makecell{FlowTrans~\cite{jin2017predicting}}   & \underline{69.38}/ \underline{69.70} /\underline{69.54} & \underline{42.90}/ \underline{52.02}/ \underline{47.02} &\underline{30.19}/ \underline{47.56}/ \underline{36.94} \\
\makecell{EgoGAN (Ours)} &{\fontseries{b}\selectfont 70.89}/ {\fontseries{b}\selectfont 71.24}/ {\fontseries{b}\selectfont 71.07} &{\fontseries{b}\selectfont 43.79}/ {\fontseries{b}\selectfont 53.23}/ {\fontseries{b}\selectfont 48.05}& {\fontseries{b}\selectfont 31.39}/ {\fontseries{b}\selectfont 48.57}/ {\fontseries{b}\selectfont 38.14}\\

\hline
\end{tabular}}}

\subfloat[Experimental Results on EGTEA Gaze+ Dataset]{
\tablestyle{8pt}{1.0}
\scalebox{0.92}{
\begin{tabular}{c|ccc}
\hline
\multicolumn{1}{c|}{\multirow{2}{*}{Method}}          
&\multicolumn{3}{c}{EGTEA (Precision/ Recall/ F1 Score)}  \\ \cline{2-4}
\multicolumn{1}{c|}{}   & short-term & middle-term  & long-term          \\ \hline 
\makecell{X2X~\cite{luc2017predicting}}   & 42.96/ 59.32/ 49.84  &38.70/ 59.89/ 47.01 & 36.55/ 59.67/ 45.33 \\
\makecell{ConvLSTM~\cite{rochan2018future}}   & \underline{44.55}/ 59.43/ 50.93 & 38.28/ {\fontseries{b}\selectfont 63.54}/ 47.78 & \underline{36.58}/ \underline{62.04}/ \underline{46.03} \\
\makecell{FlowTrans~\cite{jin2017predicting}}   & 44.22/ \underline{61.36}/ \underline{51.40} &\underline{40.38}/ 58.62/ \underline{47.82} & 35.04/ {\fontseries{b}\selectfont 64.34}/ 45.37 \\
\makecell{EgoGAN (Ours)} &{\fontseries{b}\selectfont 44.91}/ {\fontseries{b}\selectfont 61.48}/ {\fontseries{b}\selectfont 51.91}& {\fontseries{b}\selectfont 41.10}/ \underline{59.90}/ {\fontseries{b}\selectfont 48.75} & {\fontseries{b}\selectfont 38.16}/ 59.88/ {\fontseries{b}\selectfont 46.61} \\ 
\hline
\end{tabular}}}
}
\label{table:results}
\end{table*} 
\subsection{Comparison to State-of-the-Art Methods}
\label{sec:results}
We are the first to address the challenging problem of future hand segmentation from the egocentric video. We note that another branch of prior work considered the related problem of future image segmentation~\cite{yang2019video,tsai2016video,xu2018dynamic,chandra2018deep,nilsson2018semantic}, track instances masks over time, and therefore can not be used to address the future segmentation problem where the future video frames are not available as inputs for the tracking model. Therefore, we adapt previous state-of-the-art future image segmentation methods to our problem setting and consider the following strong baselines (additional discussion of the baseline choices can be found in the supplementary material):

\noindent \textbullet X2X~\cite{luc2017predicting} proposes a recursive method that uses the anticipated mask at time step $t+1$ as an input to predict the future masks at time step $t+2$, and so forth.

\noindent \textbullet FlowTrans~\cite{jin2017predicting} jointly predicts the masks and optical flow at time step $t+1$ and recursively predicts the future masks with preceding flow and masks.

\noindent \textbullet ConvLSTM~\cite{rochan2018future} uses a Convolutional LSTM to model the temporal relationships of image features, and uses both the sequence of image features and the output of the ConvLSTM module for future image segmentation.

It is worth noting that the baseline methods~\cite{luc2017predicting,jin2017predicting,rochan2018future} adopt a weaker backbone network than ours. To show that the performance gain of our method does not come from a stronger video feature encoder, we re-implement the above methods with the same I3D-Res50 backbone network as ours. Moreover, both FlowTrans and ConvLSTM assume accurate semantic segmentation of observable video frames is available as input, but our model seeks to forecast future hand segmentation using only raw video frames, and thus is a more challenging and practical setting. In addition, accurate semantic segmentation results on egocentric video frames are difficult to obtain due to the domain gap and lack of training data. Therefore, for a fair comparison, we implement the ConvLSTM and FlowTrans models to take the same input as our method. In our supplementary materials, we show that using the segmentation results from the pre-trained segmentation network as inputs will compromise the performance of FlowTrans and ConvLSTM.




The experimental results are summarized in Table~\ref{table:results}. Among all baseline methods, FlowTrans achieves the best performance for short-term anticipation. However, it is less effective for long-term anticipation, due to the error accumulation of predicted future optical flow. ConvLSTM can better capture the long-term temporal relationship and thereby achieving the best baseline performance for long-term anticipation. 
Instead of encoding the temporal connection with recursive prediction, we found that the 3D deconvolution operation is effective for capturing the temporal correlation of anticipation video segments, and in doing so, it helps capture the future hand masks in one shot. More importantly, our method outperforms previous best results (underlined in Table~\ref{table:results}) by $+1.5\%/1.0\%/1.2\%$ and $+0.5\%/0.9\%/0.6\%$ in F1 score for short/middle/long term hand mask anticipation on EPIC-Kitchens and EGTEA, respectively. Once again, these results demonstrate the benefits of explicitly modeling future head motion with a GAN. 

\begin{figure*}[!t]
\centering
\includegraphics[width=0.84\linewidth]{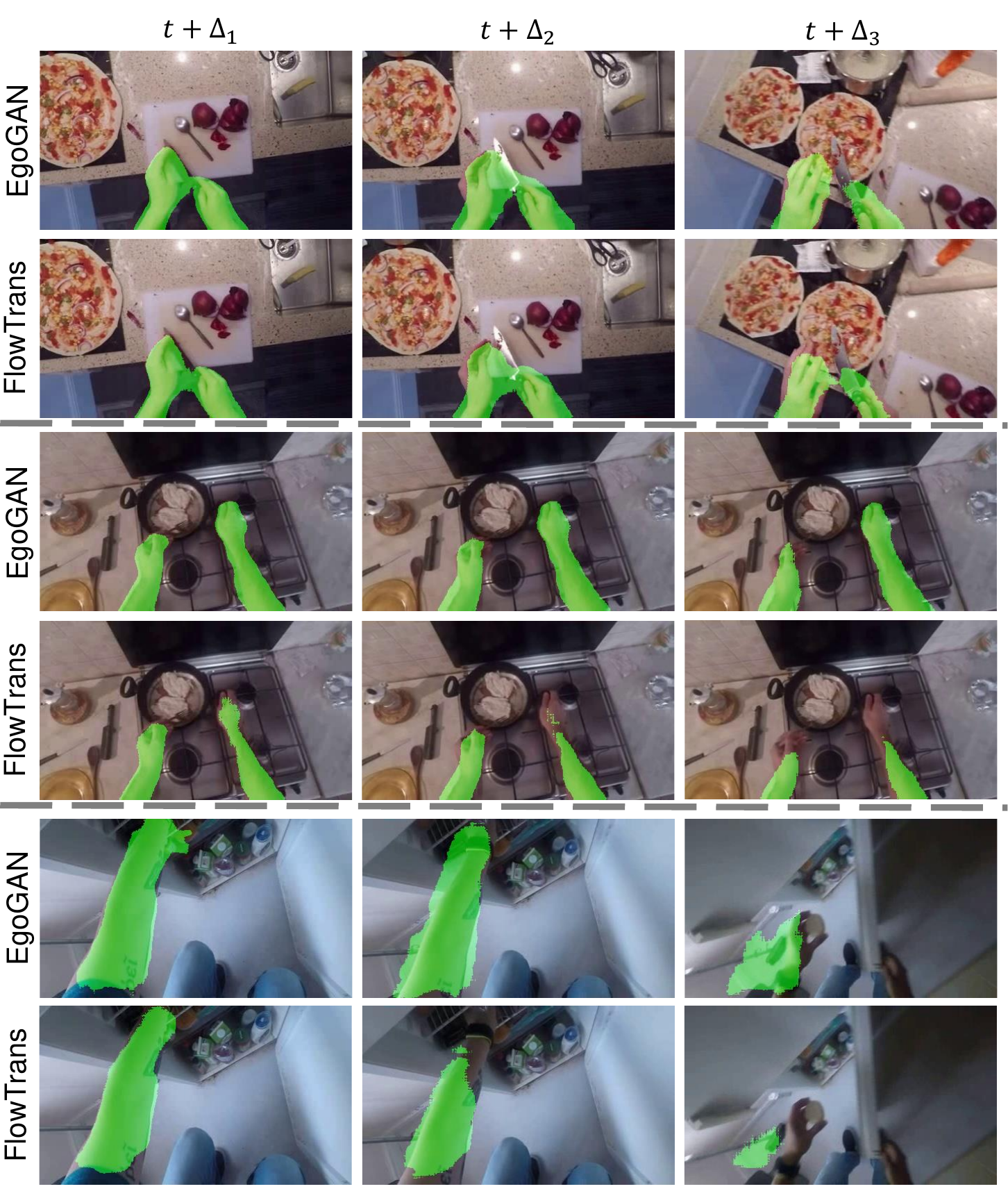}
\captionof{figure}{\textit{Visualization of our results}. From left to right, each column presents the future hand segmentation results of short-term ($t+\Delta_1$), middle-term ($t+\Delta_2$), and long-term($t+\Delta_3$) time steps from the EPIC-Kitchens dataset. Predictions from our method \textit{EgoGAN} and the best baseline \textit{FlowTrans} are presented in each sample. (See more discussion in Sec.~\ref{sec:discussion})}

\label{fig:handvis}
\end{figure*}

\subsection{Discussion}
\label{sec:discussion}

\noindent\textbf{Visualization}.\ We visualize the results from both our method \textit{EgoGAN} and the best baseline \textit{FlowTrans} on EPIC-Kitchens in Fig.~\ref{fig:handvis}. Even though our proposed problem of future hand segmentation from egocentric video poses a formidable challenge in computer vision, our method can more accurately predict the hand region of future frames compared to FlowTrans, together with capturing the hand shape and poses. Notably, as the uncertainty increases with the anticipation time, our model may produce blurry predictions, yet can still robustly localize the hand region. The video demo in our supplementary material also suggests that our approach can produce satisfying results even when there are drastic hand and head movements. We conjecture that our model can better forecast the scene context change driven by head motion, and thereby more accurately predicts future hand masks.



\noindent\textbf{Remarks}.\ To summarize, our quantitative results indicate that future head motion carries important information for future hand movements. We show that explicitly modeling the underlying distribution of possible future hand movements with a GAN enables the model to predict the future hand masks more accurately. Another important takeaway is that our method is more effective than directly using a GAN for predicting future hand masks, as reported in Table~\ref{table:ablation}. Furthermore, our visualizations demonstrate that our method can effectively predict future hand masks.

\noindent\textbf{Limitations and Future Work}.\ We also point out the limitations of our method. Since the hand mask ground truth does not differentiate left and right hands, our method cannot make separate predictions for the left and right hands. Recent work~\cite{Shan20} does have the capability of localizing left and right hand bounding boxes separately during human-object interaction, and we plan to explore this direction in our future work on visual anticipation. In addition, our work does not explicitly exploit the action and object features for future hand prediction and will leave this for our future efforts. Nonetheless, our work investigates a novel and important problem in egocentric vision, and offers insight into visual anticipation and video pixel generation.



\section{Conclusion}
In this paper, we introduce the novel task of predicting a time series of future hand masks from egocentric videos. We present a novel deep generative model EgoGAN to address our proposed problem. The key innovation of our method is to use a GAN module that explicitly models the underlying distribution of possible future head motion for a more accurate prediction of future hand masks. We demonstrate the benefits of our method on two egocentric benchmark datasets, EGTEA Gaze+ and EPIC-Kitchens 55. We believe our work provides an essential step for visual anticipation as well as video pixel generation, and points to new research directions in the egocentric video. 
\newline
\noindent \textbf{Acknowledgments}. Portions of this project were supported in part by a gift from Facebook. We thank Fiona Ryan for the valuable feedback.

\bibliographystyle{splncs04}
\bibliography{egbib}

\def\ECCVSubNumber{1710}  

\title{Supplementary Materials for: \\ ``Generative Adversarial Network for Future Hand Segmentation from Egocentric Video''} 

\titlerunning{Future Hand Segmentation}
\author{Wenqi Jia$^\star$\and
Miao Liu\thanks{Equal contribution.}\and
James M. Rehg}
\institute{Georgia Institute of Technology, Atlanta, United States}
\authorrunning{W. Jia et al.}
\maketitle
This is the supplementary material for our ECCV 2022 paper, titled ``Generative Adversarial Network for Future Hand Segmentation from Egocentric Video''. 
The contents are organized as follows.
\begin{itemize}
\item \hyperref[sec:s1]{1} Network Architecture.
\item \hyperref[sec:s2]{2} Results Using Semantic Masks as Inputs

\item \hyperref[sec:s3]{3} Results Using I3D-Res101 Backbone
\item \hyperref[sec:s4]{4} Results of Generated Future Head Motion
\item \hyperref[sec:s5]{5} Zero-Motion Baseline
\item \hyperref[sec:s6]{6} Pseudo Ground Truth of Hand Masks
\item \hyperref[sec:s7]{7} Additional Visualizations
\item \hyperref[sec:s8]{8} Code and Licenses
\end{itemize}

\section{Network Architecture}
\label{sec:s1}
We present the architecture details of the 3DFCN backbone network ($\phi_E$ and $\phi_D$), Generator Network (G), and Discriminator Network (D) in Table~\ref{table:structure}. Note that we downsample the head motion flow both spatially and temporally, resulting into a tensor with size $4\times7\times7\times2$. As shown in ID $10$ from Table~\ref{table:structure}, we directly concatenate the generated head motion flow map with the encoded video features and feed the concatenated tensors into the decoder network. Other work~\cite{jin2017predicting} proposed to combine optical flow features, learned by a convolutional operation, with video features for future segmentation. However, our empirical finding is that a convolutional flow feature extractor is not necessary, as the decoder can effectively understand the motion pattern incorporated in the low-resolution head motion flow map. 

\begin{table*}[t]
\centering
\caption{\textit{Results using semantic mask as inputs}. Our method outperforms the second-best results (across all methods) by $1.3\%$ on EPIC-Kitchens in average F1 score. The best results are highlighted with \textbf{boldface}, and the second-best results are \underline{underlined}.}
\label{table:seg}
\centering
{
\tablestyle{6pt}{1.0}
\begin{tabular}{c|ccc}
\hline
\multicolumn{1}{c|}{\multirow{2}{*}{Method}}           &\multicolumn{3}{c}{EPIC-Kitchens (Precision/ Recall/ F1 Score)} \\ \cline{2-4}
\multicolumn{1}{c|}{} & short-term & middle-term  & long-term     \\ \hline
\makecell{S2S}       & 24.85/ {\fontseries{b}\selectfont 56.12}/ 34.45 & 27.59/ {\fontseries{b}\selectfont 54.69}/ 36.68 & 26.86/ {\fontseries{b}\selectfont 52.59}/ 35.55 \\
\makecell{ConvLSTM}  & \underline{28.24}/ 45.48/ 34.84 & \underline{30.20}/ 49.00/ 37.37 & \underline{29.45}/ 47.70/ 36.42\\
\makecell{FlowTrans} & 27.97/ 47.82/ \underline{35.30} & 29.58/ \underline{52.07}/ \underline{37.73} & 28.95/ \underline{49.89}/ \underline{36.64}\\
\makecell{Ours}      & {\fontseries{b}\selectfont 29.09}/ \underline{47.86}/ {\fontseries{b}\selectfont 36.19} & {\fontseries{b}\selectfont 33.14}/ 47.50/ {\fontseries{b}\selectfont 39.04} & {\fontseries{b}\selectfont 32.68}/ 45.05/ {\fontseries{b}\selectfont 37.88}\\
\hline
\end{tabular}}
\end{table*}

\section{Results Using Semantic Masks as Inputs}
\label{sec:s2}
As discussed in Sec.4.3 from our main paper, the original ConvLSTM~\cite{rochan2018future} and FlowTrans~\cite{jin2017predicting} take accurate semantic segmentation masks, from a fine-tuned semantic segmentation model, as inputs. However,  the semantic segmentation annotation is not available on the existing egocentric video datasets~\cite{li2020eye,Damen2018EPICKITCHENS} to fine-tune a segmentation model. And the pre-trained Mask-RCNN model obtains sub-optimal results on egocentric video datasets~\cite{damen2021rescaling}. In this section, we present the results of ConvLSTM, FlowTrans, S2S \footnote{X2X model described in our main paper was denoted as S2S in ~\cite{luc2017predicting}, when using the semantic mask as inputs}, as well as our method using the semantic segmentation from~\cite{damen2021rescaling} as inputs. The experimental results are summarized in Table~\ref{table:seg}. Notably, using the semantic segmentation results from a pre-trained model decreases the model performance on short-term and middle-term hand mask anticipation, yet slightly improves the long-term future hand segmentation results for all methods. More importantly, when using the semantic masks as inputs, our model outperforms the second-best results (across all methods) by $0.9\%/1.3\%/1.2\%$ in F1 Score for short/middle/long-term future hand segmentation. These results further demonstrate the robustness of our method. It is worth noting that another relevant work from~\cite{chiu2020segmenting} also addresses the future segmentation problem, and can adopt either raw video frames or semantic masks as inputs. However, the official implementation of~\cite{chiu2020segmenting} is under construction. And we found the training of our implementation of~\cite{chiu2020segmenting} to be unstable under the egocentric setting, probably because that the drastic change of scene context incurs additional barriers for the distillation model to generalize.

\section{Results Using I3D-Res101 Backbone}
\label{sec:s3}
We further show our method can generalize to different backbone encoder networks. In Table~\ref{table:res101}, we report the future hand segmentation results of both our method and 3DFCN baseline using I3DRes50 and I3DRes101 backbone. As discussed in the main paper, the performance improvement of our method (EgoGAN-I3DRes50 vs 3DFCN-I3DRes50) is larger than adopting a denser backbone model (3DFCN-I3DRes101 vs 3DFCN-Res50). Moreover, the EgoGAN model with I3D-Res101 improves 3DFCN-I3DRes101 by $+0.1\%/0.1\%/0.3\%$ on EGTEA and $+0.5\%/0.4\%/0.7\%$ on EPIC-Kitchens. These results further show the robustness of our method. (Note that the performance improvement on EGTEA is relatively small with I3DRes101 backbone, due to the limited training data and dense backbone encoder.) 

\begin{table}[!t]
\centering
\caption{\textit{Experimental results using different backbone networks}. Our model achieves consistent performance improvement when using different backbone networks. (See more discussion in Sec.~\ref{sec:s4})}
{
\subfloat[Experimental Results on EPIC-Kitchens Dataset]{
\tablestyle{5pt}{1.0}
\scalebox{0.91}{
\begin{tabular}{c|c|ccc}
\hline
\multicolumn{1}{c|}{\multirow{2}{*}{Method}}&\multicolumn{1}{c|}{\multirow{2}{*}{Backbone}}&\multicolumn{3}{c}{Epic-Kitchens (Precision/ Recall/ F1 Score)} \\ \cline{3-5}
\multicolumn{1}{c|}{} & & short-term & middle-term  & long-term     \\ \hline 
\multicolumn{1}{c|}{\multirow{2}{*}{3DFCN}}  & I3DRes50 & 69.51/ 70.81/ 70.15 & 42.51/ 51.66/ 46.64 & 29.88/ 47.46/ 36.67 \\
& I3DRes101 & 69.48/ 70.96/ 70.21 & 42.32/ 52.80/ 46.98 & 29.97/ 48.37/ 37.01 \\ \hline 
\multicolumn{1}{c|}{\multirow{2}{*}{EgoGAN}} & I3DRes50 & 70.89/ 71.24/ 71.07 & 43.79/ 53.23/ 48.05 & 31.39/ 48.57/ 38.14\\
& I3DRes101 & 69.17/ 74.05/ 71.53 & 44.09/ 53.79/ 48.46 & 30.79/ 52.60/ 38.85 \\
\hline
\end{tabular}}}

\subfloat[Experimental Results on EGTEA Gaze+ Dataset]{
\tablestyle{5pt}{1.0}
\scalebox{0.9}{
\begin{tabular}{c|c|ccc}
\hline
\multicolumn{1}{c|}{\multirow{2}{*}{Method}}&\multicolumn{1}{c|}{\multirow{2}{*}{Backbone}}&\multicolumn{3}{c}{EGTEA (Precision/ Recall/ F1 Score)} \\ \cline{3-5}
\multicolumn{1}{c|}{}    & & short-term & middle-term  & long-term     \\ \hline 
\multicolumn{1}{c|}{\multirow{2}{*}{3DFCN}}  & I3DRes50 & 43.62/ 61.69/ 51.11 &40.25/ 58.93/ 47.83 & 37.83/ 58.32/ 45.89 \\
& I3DRes101 & 44.66/ 61.81/ 51.85 & 40.49/ 59.72/ 48.26 & 35.70/ 66.18/ 46.38 \\ \hline 
\multicolumn{1}{c|}{\multirow{2}{*}{EgoGAN}} & I3DRes50 & 44.91/ 61.48/ 51.91 & 41.10/ 59.90/ 48.75 & 38.16/ 59.88/ 46.61 \\ 
& I3DRes101 & 45.69/ 60.42/ 52.03 & 39.40/ 64.27/ 48.85 & 36.92/ 64.43/ 46.94 \\
\hline
\end{tabular}}}
}
\label{table:res101}
\end{table}

\begin{table}[h]
\caption{\textit{Experimental results on generated future head motion}. We calculate the endpoint error (EPE) between the generated head motion and the ground truth head motion. Our method outperforms HeadReg on the EPIC-Kitchens dataset and works on-par with HeadReg on the EGTEA dataset.}
\centering
\setlength{\tabcolsep}{6pt} 
\renewcommand{\arraystretch}{1} 
\footnotesize
\begin{tabular}{c|cc}
\hline
Method       & Epic-Kitchens (EPE $\downarrow$) & EGTEA (EPE $\downarrow$) \\ \hline 
HeadReg      & 10.39 & 5.27 \\
EgoGAN(Ours) & 7.08  & 5.16 \\
\hline
\end{tabular}

\label{table:head}
\end{table}

\section{Results of Generated Future Head Motion}
\label{sec:s4}
Our model also has the capability of generating future head motions. In Table~\ref{table:head}, we compare our methods with HeadReg -- the only baseline model that predicts future head motion. We use the standard endpoint error (EPE) as evaluation metric. On the EPIC-Kitchens dataset, our method outperforms HeadReg by a significant margin. The performance improvement of our method is smaller on the EGTEA dataset, due to fewer available training samples. These results suggest that the GAN from our model can generates more realistic future head motion.

\section{Zero-Motion Baseline}
\label{sec:s5}
We conduct additional experiments to show our method does not simply generate a trivial solution that predicts an ``average'' hand mask shared across all time steps. Specifically, we consider a zero-motion baseline model that has the same model design as our method, yet ignores the hand motion. Therefore, the future hand segmentation results $h^{t+\Delta_2}$ and $h^{t+\Delta_3}$ are identical to $h^{t+\Delta_1}$. This baseline model achieves a F1 score of $46.14\%$ and $34.30\%$ for middle-term and long-term future hand segmentation on EPIC-Kitchens, which lags behind our full model ($1.9\%/3.8\%\downarrow$). These results further demonstrate that our method is capable of capturing meaningful hand movements.

\begin{figure*}[!ht]
\centering
\includegraphics[width=0.97\linewidth]{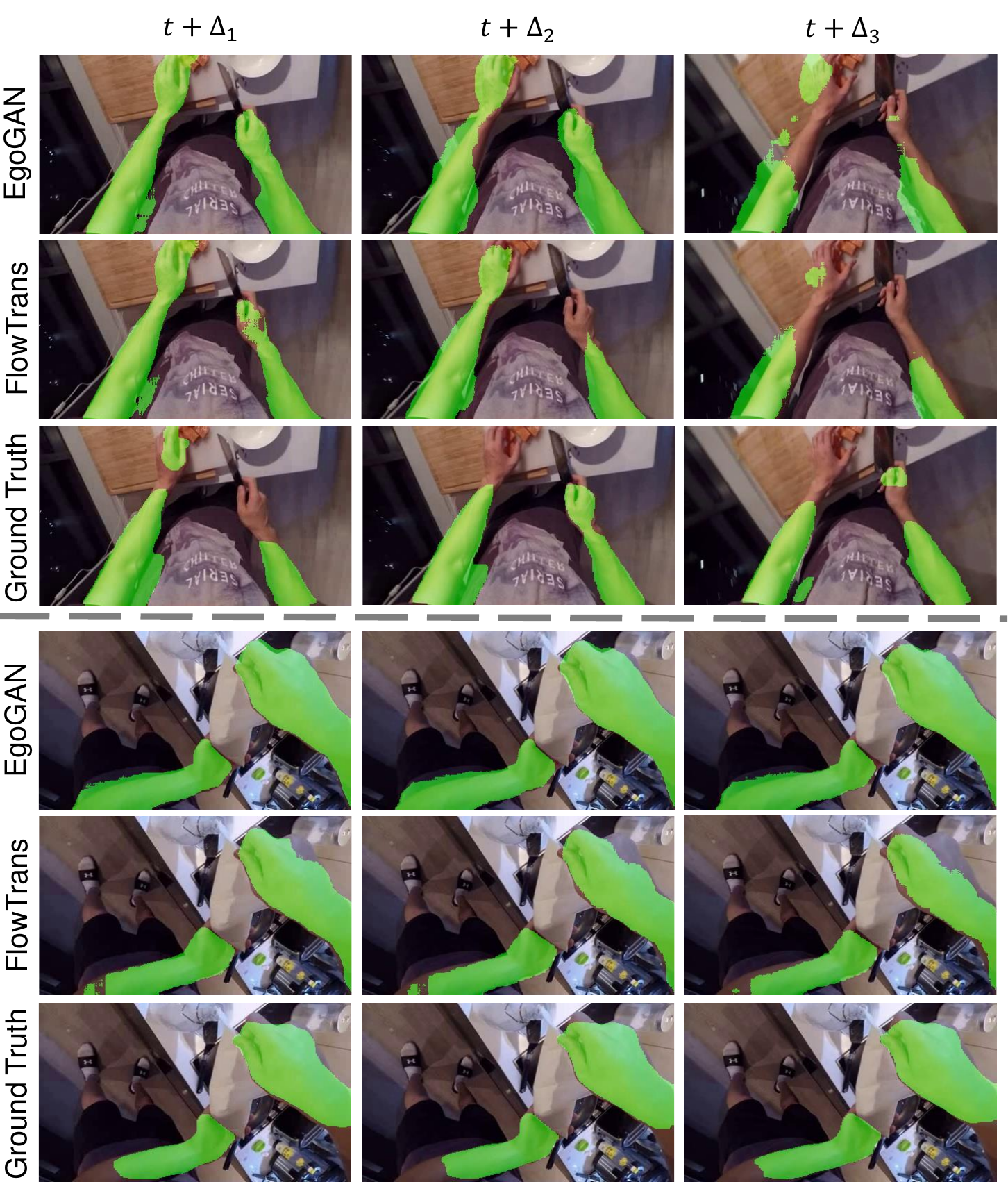}
\captionof{figure}{\textit{Visualization of hand mask ground truth, and prediction results from our model and FlowTrans}. Because of the sub-optimal ground truth, the quantitative results can not demonstrate the true performance improvement of our approach. (See more discussion in Sec.~\ref{sec:s6})}
\label{fig:gtpred}
\end{figure*}

\section{Pseudo Ground Truth of Hand Masks}
\label{sec:s6}
Though the domain adaption method from~\cite{Cai_2020_CVPR} can generate high quality hand segmentation results, they method still yields to the challenging factor of egocentric video, and thus may produce sub-optimal hand segmentation results. Therefore our quantitative experiments cannot reflect the true performance improvement of our method. In Fig.~\ref{fig:gtpred}, we visualize, hand masks ground truth and prediction results from both our method and FlowTrans. Even though our method demonstrate stronger generalizing capability and predicts more detailed hand shapes, our model has lower F1 score than FlowTrans due to the inaccurate hand masks ground truth.

\begin{figure*}
\centering
\includegraphics[width=0.97\linewidth]{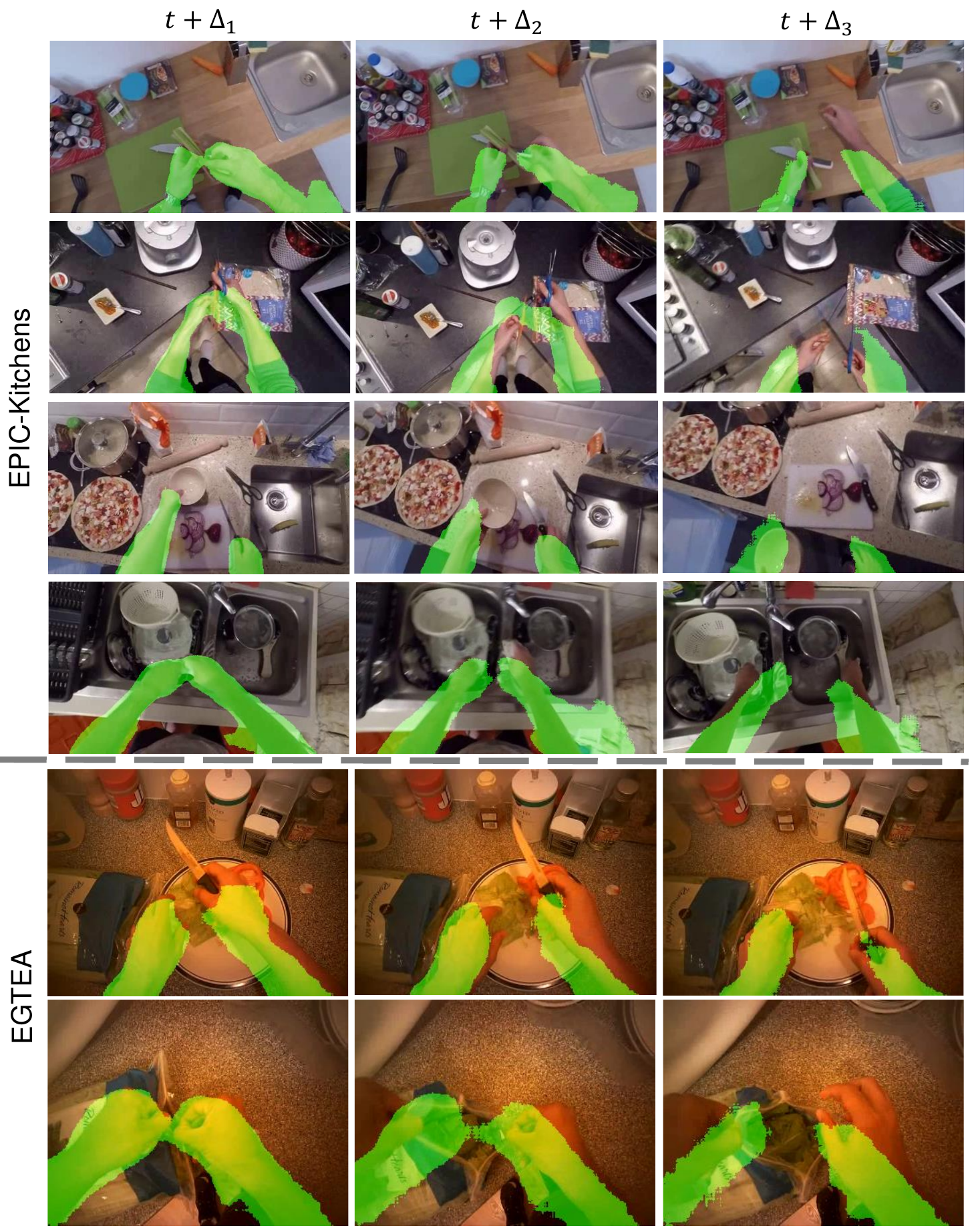}
\captionof{figure}{\textit{Additional Visualization of predicted future hand masks}.  From left to right, each column presents the RGB frame at time step $t$ and the short-term ($t+\Delta_1$), middle-term ($t+\Delta_2$), and long-term($t+\Delta_3$) future hand segmentation results. The first four rows are our results on the EPIC-Kitchens dataset, and the rest three rows are our results on the EGTEA dataset. }
\label{fig:morehandvis}
\end{figure*}

\section{Additional Visualizations}
\label{sec:s7}
We provide additional visualizations of our results in Fig.~\ref{fig:morehandvis}. Our method can effectively predict future hand masks. However, the model performance drops as the anticipation time increases. This is the same pitfall shared by previous works~\cite{liu2019forecasting} on visual anticipation. Note that we also provide the video demos of our method.

\section{Code and Licenses}
\label{sec:s8}
The usage of the EPIC-Kitchens Dataset is under the Attribution-NonCommercial 4.0 International License\footnote{\url{https://creativecommons.org/licenses/by-nc/4.0/}}. EGTEA Gaze+ dataset did not provide a license but can be used for research purposes. Our implementation is built on top of~\cite{fan2020pyslowfast}, which is under the Apache License\footnote{\url{https://github.com/facebookresearch/SlowFast/blob/main/LICENSE}}. Our code will be available at \url{https://github.com/VJWQ/EgoGAN.git}. 

\begin{table*}[!t]
\small
\def\arraystretch{1.36}
\setlength{\tabcolsep}{0.3pt}
\centering
\caption{\textit{Network architecture of our EgoGAN}. We omit the residual connection in backbone I3D-Res50 for simplification. And we present the tensor dimension during training time.}

\scalebox{0.71}{
\begin{tabular}{c|c|c|c|c|c|c}
\hline 
\multirow{2}{*}{\textbf{ID}} & \multirow{2}{*}{\textbf{Branch}} & \multirow{2}{*}{\textbf{Type}} & \multirow{2}{*}{\begin{tabular}[c]{@{}c@{}}\textbf{Kernel Size}\\ THW,(C)\end{tabular}} & \multirow{2}{*}{\begin{tabular}[c]{@{}c@{}}\textbf{Stride}\\ THW\end{tabular}} & \multirow{2}{*}{\begin{tabular}[c]{@{}c@{}}\textbf{Output Size}\\ THWC\end{tabular}} & \multirow{2}{*}{\textbf{Comments}} \\ &   &   &   &   &   & \\ \hline 
1   & \multirow{24}{*}{\begin{tabular}[c]{@{}c@{}}Encoder\\ Input Size:\\ $8\times224\times224\times3$\end{tabular}} 
& Conv3D    & $5\times7\times7,64$ & $1\times2\times2$ & $8\times112\times112\times64$ \\ \cline{3-7}
2  & & MaxPool1  & $1\times3\times3$  & $1\times2\times2$ & $8\times56\times56\times64$ & Skip connect with 24 \\ \cline{3-7}
3  & & \begin{tabular}[c]{@{}c@{}}Layer1\\Bottleneck 0-2 \end{tabular}    & \begin{tabular}[c]{@{}c@{}}$3\times1\times1,64$\\ $1\times3\times3,64$\\ $1\times1\times1, 256$\end{tabular} ($\times3$)     & \begin{tabular}[c]{@{}c@{}}$1\times1\times1$\\ $1\times1\times1$\\$1\times1\times1$\end{tabular} ($\times3$)     & $8\times56\times56\times256$ &  \\ \cline{3-7}
4  & & MaxPool2 & $2\times1\times1$ & $2\times1\times1$ & $4\times56\times56\times256$ & Skip connect with 23 \\ \cline{3-7}
5  & & \begin{tabular}[c]{@{}c@{}}Layer2\\Bottleneck 0 \end{tabular} & \begin{tabular}[c]{@{}c@{}}$3\times1\times1, 128$\\ $1\times3\times3, 128$\\$1\times1\times1, 512$\end{tabular} & \begin{tabular}[c]{@{}c@{}}$1\times1\times1$\\ $1\times2\times2$\\$1\times1\times1$\end{tabular} & & \\ \cline{3-7}
6  & & \begin{tabular}[c]{@{}c@{}}Layer2\\Bottleneck 1-3 \end{tabular} & \begin{tabular}[c]{@{}c@{}}$3\times1\times1, 128$\\ $1\times3\times3, 128$\\$1\times1\times1, 512$\end{tabular}  ($\times3$) & \begin{tabular}[c]{@{}c@{}}$1\times1\times1$\\ $1\times2\times2$\\$1\times1\times1$\end{tabular} ($\times3$) & $4\times28\times28\times512$ & \begin{tabular}[c]{@{}c@{}} \\ \end{tabular}  Skip connect with 22 \\ \cline{3-7}
7  & & \begin{tabular}[c]{@{}c@{}}Layer3\\Bottleneck 0 \end{tabular} & \begin{tabular}[c]{@{}c@{}}$3\times1\times1, 256$\\ $1\times3\times3, 256$\\$1\times1\times1, 1024$\end{tabular} & \begin{tabular}[c]{@{}c@{}}$1\times1\times1$\\ $1\times2\times2$\\$1\times1\times1$\end{tabular} &  &  \\ \cline{3-7}
8  & & \begin{tabular}[c]{@{}c@{}}Layer3\\Bottleneck 1-5 \end{tabular}   & \begin{tabular}[c]{@{}c@{}}$3\times1\times1, 256$\\ $1\times3\times3, 256$\\$1\times1\times1, 1024$\end{tabular}  ($\times5$) & \begin{tabular}[c]{@{}c@{}}$1\times1\times1$\\ $1\times1\times1$\\$1\times1\times1$\end{tabular} ($\times5$) & $4\times14\times14\times1024$ &  Skip connect with 21 \\ \cline{3-7}
9  & & \begin{tabular}[c]{@{}c@{}}Layer4\\Bottleneck 0 \end{tabular} & \begin{tabular}[c]{@{}c@{}}$3\times1\times1, 128$\\ $1\times3\times3, 128$\\$1\times1\times1, 512$\end{tabular}     & \begin{tabular}[c]{@{}c@{}}$1\times1\times1$\\ $1\times2\times2$\\$1\times1\times1$\end{tabular}     &  &  \\ \cline{3-7}
10 & & \begin{tabular}[c]{@{}c@{}}Layer4\\Bottleneck 1-2 \end{tabular}    & \begin{tabular}[c]{@{}c@{}}$3\times1\times1, 128$\\ $1\times3\times3, 128$\\$1\times1\times1, 512$\end{tabular}  ($\times2$) & \begin{tabular}[c]{@{}c@{}}$1\times1\times1$\\ $1\times2\times2$\\$1\times1\times1$\end{tabular} ($\times2$) & $4\times7\times7\times2048$ & \begin{tabular}[c]{@{}c@{}} \\ \end{tabular} \begin{tabular}[c]{@{}c@{}} Concat with  generated \\ future head motion  \end{tabular} \\ \cline{3-7} \hline
11 & \multirow{5}{*}{\begin{tabular}[c]{@{}c@{}}Generator \\ Network (G)\\ Input Size:\\ $4\times7\times7\times2048$ \end{tabular}} & Conv3d 1 & $1\times1\times1, 2048$ & $1\times1\times1$ & $4\times7\times7\times1024$ & \\ \cline{3-7}
12 & & Conv3d 2 & $1\times1\times1, 1024$ & $1\times1\times1$ & $4\times7\times7\times512$ & \\ \cline{3-7}
13 & & Conv3d 3 & $1\times1\times1, 512$ & $1\times1\times1$ & $4\times7\times7\times2$ & \\ \cline{3-7}
14 & & \begin{tabular}[c]{@{}c@{}}Tanh \end{tabular} &  &  &$4\times7\times7\times2$  & \begin{tabular}[c]{@{}c@{}}Input for Decoder \\ \& Discriminator \end{tabular}\\ \hline
15 & \multirow{4}{*}{\begin{tabular}[c]{@{}c@{}}\\Discriminator \\ Network (D)\\ Input Size:\\ $4\times7\times7\times2$ \end{tabular}} & Conv3d 1 & $1\times3\times3, 2$ & $1\times1\times1$ & $4\times5\times5\times32$ & \\ \cline{3-7}
16 & & Conv3d 2 & $1\times3\times3, 32$ & $1\times1\times1$ & $4\times3\times3\times64$ & \\ \cline{3-7}
17 & & Conv3d 3 & $1\times3\times3, 64$ & $1\times1\times1$ & $2\times1\times1\times128$ & \\ \cline{3-7}
18 & & Adaptive Avg Pooling & & & $1\times1\times1\times128$ &  \\ \cline{3-7}
19 & & Linear & & & $1$ & \\ \cline{3-7}
20 & & Sigmoid & & & $1$ & BCE loss \\ \hline

21 & \multirow{9}{*}{\begin{tabular}[c]{@{}c@{}}Decoder\\ Input Size:\\ $4\times7\times7\times2050$ \end{tabular}}
   & ConvTranspose3d 1 & $1\times3\times3,2050$ & $1\times2\times2$ & $4\times14\times14\times1024$ & Skip connect with 8 \\ \cline{3-7}
22 & & ConvTranspose3d 2 & $1\times3\times3,1024$ & $1\times2\times2$ & $4\times28\times28\times512$ & Skip connect with 6 \\ \cline{3-7}
23 & & ConvTranspose3d 3 & $1\times3\times3,512$ & $1\times2\times2$ & $4\times56\times56\times256$ & Skip connect with 4 \\ \cline{3-7}
24 & & ConvTranspose3d 4 & $3\times3\times3,256$ & $1\times1\times1$ & $8\times56\times56\times64$ & Skip connect with 2 \\ \cline{3-7}
25 & & ConvTranspose3d 5 & $1\times5\times5,64$ & $1\times4\times4$ & $8\times224\times224\times64$ & \\ \cline{3-7} 
26 & & Conv3d 1 & $4\times1\times1,64$ & $1\times1\times1$ & $5\times224\times224\times32$ & \\ \cline{3-7} 
27 & & Conv3d 2 & $3\times1\times1,32$ & $1\times1\times1$ & $3\times224\times224\times16$ & \\ \cline{3-7}
28 & & \begin{tabular}[c]{@{}c@{}}Conv3d 3\\(Classifier) \end{tabular} & $1\times1\times1,16$ & $1\times1\times1$ & $3\times224\times224\times1$  & BCE loss \\ \cline{3-7}
\hline
\end{tabular}}
\label{table:structure}
\end{table*}

\clearpage

\end{document}